\documentclass{article}
 \pdfoutput=1
\usepackage{microtype}
\usepackage{graphicx}
\usepackage{subfigure}
\usepackage{booktabs} 

\usepackage{hyperref}

\usepackage[accepted]{icml2021}
\usepackage{amsmath}
\usepackage{amssymb}
\usepackage{booktabs} 
\usepackage{appendix}
\usepackage{amsthm}
\usepackage{multirow} 

\icmltitlerunning{LARNet: Lie Algebra Residual Network for Face Recognition}
\begin{document}

\twocolumn[
\icmltitle{LARNet: Lie Algebra Residual Network for Face Recognition}



\icmlsetsymbol{equal}{*}

\begin{icmlauthorlist}
\icmlauthor{Xiaolong Yang}{amss,ucas,ten}
\icmlauthor{Xiaohong Jia}{amss,ucas}
\icmlauthor{Dihong Gong}{ten}
\icmlauthor{Dong-Ming Yan}{ia,ucas}
\icmlauthor{Zhifeng Li}{ten}
\icmlauthor{Wei Liu}{ten}
\end{icmlauthorlist}

\icmlaffiliation{amss}{Academy of Mathematics and Systems Science, Chinese Academy of Sciences, Beijing, P.R.China}
\icmlaffiliation{ucas}{University of Chinese Academy of Sciences, Beijing, P.R.China}
\icmlaffiliation{ten}{Tencent Data Platform, P.R.China}
\icmlaffiliation{ia}{NLPR, Institute of Automation, Chinese Academy of Sciences, Beijing, P.R.China}

\icmlcorrespondingauthor{Xiaohong Jia, Zhifeng Li, and Wei Liu}{xhjia@amss.ac.cn; michaelzfli@tencent.com; wl2223@columbia.edu} 

\icmlkeywords{Machine Learning, ICML}

\vskip 0.3in
]



\printAffiliationsAndNotice{}  

\begin{abstract}
Face recognition is an important yet challenging problem in computer vision. A major challenge in practical face recognition applications lies in significant variations between profile and frontal faces. Traditional techniques address this challenge either by synthesizing frontal faces or by pose invariant learning. In this paper, we propose a novel method with Lie algebra theory to explore how face rotation in the 3D space affects the deep feature generation process of convolutional neural networks (CNNs). We prove that face rotation in the image space is equivalent to an additive residual component in the feature space of CNNs, which is determined solely by the rotation. Based on this theoretical finding, we further design a Lie Algebraic Residual Network (LARNet) for tackling pose robust face recognition. Our LARNet consists of a residual subnet for decoding rotation information from input face images, and a gating subnet to learn rotation magnitude for controlling the strength of the residual component contributing to the feature learning process. Comprehensive experimental evaluations on both frontal-profile face datasets and general face recognition datasets convincingly demonstrate that our method consistently outperforms the state-of-the-art ones.
\end{abstract}

\section{Introduction}\label{sec:intro}
The recent development of deep learning models and an increasing variety of datasets have greatly advanced face recognition technologies ~\cite{Liu2017,Wang2018,Deng2019}. Although  many deep  learning models are strong and robust to  face recognition conducted in unconstrained environments, there remain quite a lot of challenges for recognizing faces across different age levels ~\cite{gong2013hidden,wang2019decorrelated,wang2018orthogonal,gong2015maximum,li2016aging},  different modalities~\cite{li2014common,gong2017heterogeneous,li2016mutual,gong2013multi,Luo2021}, different poses~\cite{huang2000pose,Cao2018,Masi2016,AbdA2016}, and occlusions~\cite{song2019occlusion,zhang2007local}. In this paper, we develop a robust recognition algorithm to address the challenges for general face recognition, with a particular effect on matching faces across different poses (e.g., frontal vs. profile). Since the generalization ability of the deep model is closely related to the size of the training data, given an uneven and insufficient distribution of frontal and profile face images, the deep features tend to focus on  frontal faces, and the learning results are only biased incomplete statistics. In order to tackle this problem, some work has reconstructed more datasets by different data augmentation methods. A typical way is to enrich  input sources either by the synthesis of profile faces with appearance variations ~\cite{Masi2016b} or by a set of images as one image input~\cite{Xie2018}, so  that the need for profile data is alleviated. Another way is to combine more data information, including multi-task learning~\cite{AbdA2016,Masi2016,Yin2017} and template adaptation~\cite{Hassner2016,Cross2018}. Here, multi-task learning focuses on pose-aware targets, combined with richer information such as illumination, expression, gender, and age, to comprehensively boost the recognition performance; while the method based on template adaptation learning always creates a mean 3D model face, and by means of migration and mapping it avoids processing the 3D transformation at the image level. Nevertheless, these strategies tend to increase the unnecessary computational burden. Some other approaches use profile faces to synthesize frontal faces so that they can avoid large pose variations~\cite{Tran2017,Wang2017b,Yin2017b,Zhou2020}. However, these methods would suffer from artifacts caused by occlusions and non-rigid expressions. 

The above mentioned work mostly relies on additional data sources or additional labels. A recent new work called Deep Residual Equivalent Mapping (DREAM)~\cite{Cao2018} has further discussed the gap between those features of frontal-profile pairs simply by approximating the difference using a learning model. Their approach of exploring this gap is similar to generative adversarial network (GAN), which makes the target sample (frontal face feature)  and the generated sample (profile face feature) as close as possible through encoding and decoding. 

We observe a natural fact that  frontal-profile pairs are generated by head rotations, which should not be ignored in profile face recognition. However, rotation matrices are not easy to be embedded in CNNs. This is because the group of rotation matrices is closed under multiplication but not closed under addition, while
the addition operation appears frequently in all gradient descent calculations. Benefiting from the pose estimation work~\cite{Tuzel2008} in the field of \emph{simultaneous localization and mapping}~(SLAM), we develop a novel approach of using Lie algebra to achieve the updates  of rotation matrices in CNNs.

\begin{figure}[t]
  \centering
  \includegraphics[width=\linewidth]{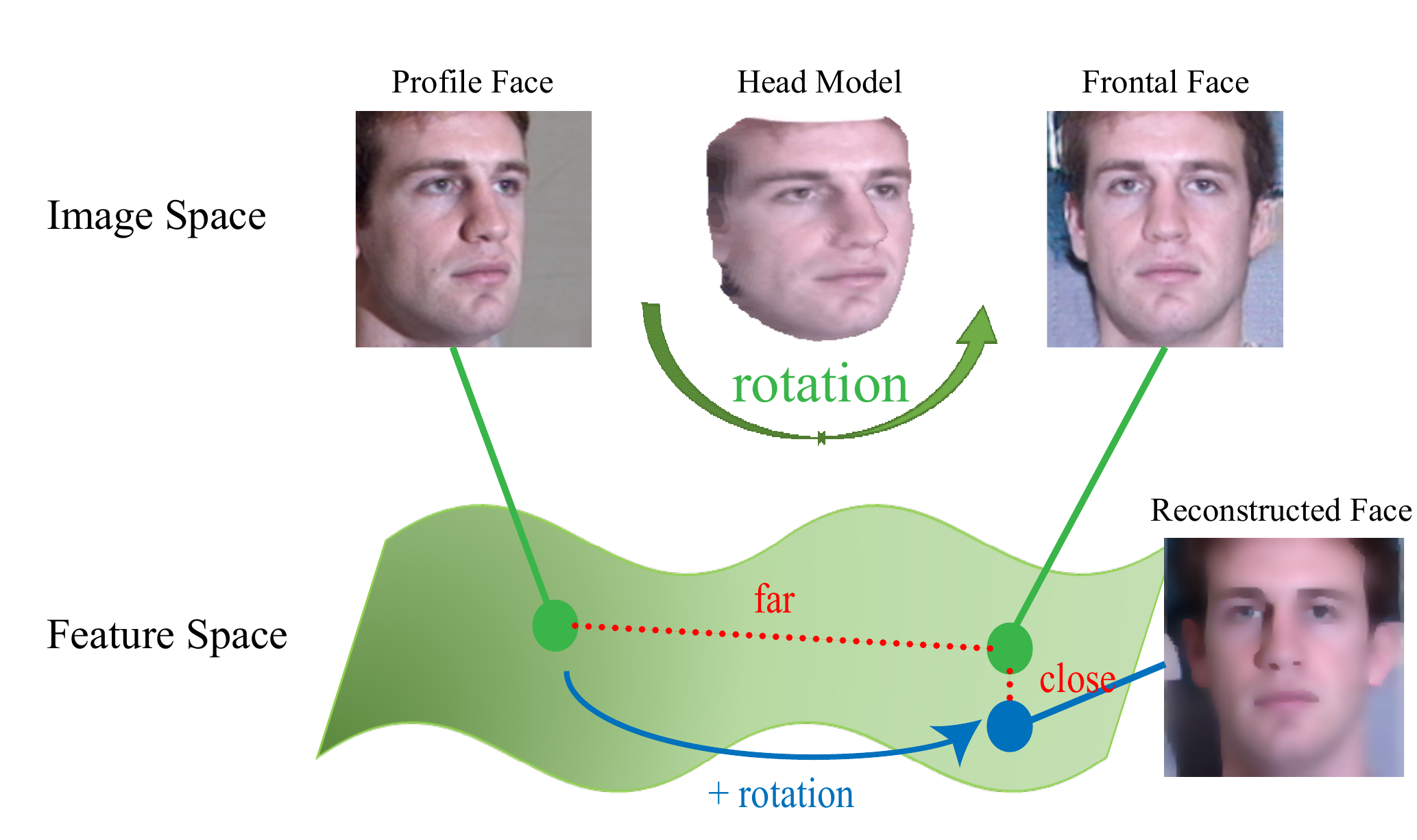}
  \vskip -3mm
  \caption{  \emph{Frontalization} or \emph{rotation} in the feature space. Naturally in the real world, a frontal-profile pair is generated by head rotation,  and we prove that a face rotation in the image space is equivalent to an additive residual component in the deep feature space. To show the equivalence, we reconstruct the image corresponding to the modified feature (blue dot) and provide the visual result for the expected frontal face. }
  \label{fig:intro}
  \vskip -3mm
\end{figure}

We prove that for each frontal-profile pair linked by a rotation, their corresponding deep features also preserve a corresponding rotation relationship. To the best of our knowledge, this is the first attempt to theoretically explore and explain the physical relationship between the features of a frontal face and its profile counterpart.  Based on this theoretical result, we propose the \emph{Lie Algebra Residual Network} (LARNet). LARNet achieves face frontalization or rotation-and-render in the feature space, as shown in Fig.~\ref{fig:intro}.  Meanwhile, we conduct comparative experiments with more than 30 solutions under various evaluation criteria and metrics,  while our method outperforms representative state-of-the-art competitors. In summary, our contributions are three-fold:
\vskip -3mm
\begin{enumerate}
\vskip -3mm
\item We theoretically prove that the features of frontal-profile pairs have a physical relationship based on rotations in a residual network using Lie algebra, which is equivalent to an additive residual component in the deep feature space of CNNs.
\vskip -3mm
\item We design a novel gating subnet based on the proof, which neither needs to modify the original backbone network structure nor relies on  a large number of modules, but brings a great performance improvement.
\vskip -3mm
\item LARNet enhances the deep model's ability in feature representation and feature classification,  and accomplishes superior performance on various datasets and in various evaluation criteria, including frontal-profile face verification-identification  and general face recognition tasks.
\vskip -3mm
\end{enumerate}

\section{Related Work}\label{sec:relawork}
We briefly discuss the most related  work of profile face recognition and large-pose face recognition. 

\noindent{\textbf{Insufficient Dataset.}} Many methods try to solve the profile face recognition problem by avoiding the unevenness of datasets. Masi~\emph{et al.}~\yrcite{Masi2016b} proposed domain-specific data augmentation with increasing training data sizes for face recognition systems, and focused on important facial appearance variations. Multicolumn Network~\cite{Xie2018} and Neural Aggregation Network (NAN) ~\cite{Yang2017} propose to use more information, such as a set of images  or videos as input,  to tackle the potential shortcomings of merely using a single image. These methods are not easy to avoid falsely matching profile faces of different identities or missing frontal and profile faces of the same identity.
\noindent{\textbf{Pose Variation.}} Many existing methods have conducted in-depth researches on large poses. Template-adaptation-based work~\cite{Hassner2016, Cross2018}  is mainly engaged with transfer learning by means of a constructed classifier and synthesizer, and pooling based on image quality and head poses. As opposed to those techniques which expect a template model to learn pose invariance,  pose aware deep learning methods~\cite{AbdA2016, Masi2016} use multiple pose-specific models and rendered face images, which reduce the sensitivity to pose variations.  More work favors using more labels instead of pose itself. Multi-task learning (MTL) has been widely used, which consists of pose, illumination, and expression estimations. Yin~\emph{et al.}~\yrcite{Yin2017b} proposed a pose-directed multi-task CNN and united the balance between different tasks. DebFace~\cite{Zhou2020} (de-biasing adversarial network) additionally takes gender, age, and race into consideration, and minimizes the correlation among feature factors so as to abate the bias influence from the other factors. These methods are effective; however, the use of multiple models and tasks tends to increase the computational cost, and the accuracy of their results is confined.

\noindent{\textbf{Frontalization.}} Since profile and large-pose faces bring more challenges, some methods directly use an available dataset to synthesize frontal faces to perform face recognition. Due to the widespread use of GAN, FF-GAN~\cite{Yin2017} and DR-GAN~\cite{Tran2017} surpass the performances of many competitors,  with disentangled encoder-decoder structure towards learning a generative and discriminative representation. With the rapid progress of 3D face reconstruction technology, projecting rendering of  frontal faces after reconstruction has also risen. Rotate-and-Render~\cite{Zhou2020} is a representative work from single-view images, and can leverage the recent advances in 3D face modeling and high-resolution GAN to constitute  building blocks, since the 3D rotation-and-render of faces can be applied to arbitrary angles without losing facial details. Note that the reconstruction and  synthesis approaches have their advantages in visualization performance, but their feature representation capability is inadequate for applications in unconstrained environments.

\noindent{\textbf{Feature Space.}} Some work considered features rather than image itself.  Shi~\emph{et al.}~\yrcite{Shi2019} proposed \emph{Probabilistic Face Embeddings} (PFEs), which represent each face image as a Gaussian distribution in the latent space. Feature Transfer Learning~\cite{Yin2019} encourages the under-represented distribution to be closer to the regular distribution. Both of these two approaches target at making the sample distribution close to a Gaussian prior.  Although in the above work features have been paid attention, specific datasets or face recognition in the real world cannot yet guarantee that samples are in a Gaussian distribution. Another representative work  is DREAM~\cite{Cao2018},  which uses a residual network to directly modify the features of a profile face to the frontal one. DREAM  bridges frontal-profile feature pairs by a mapping learned by deep learning; however, their ad-hoc designed results reach a bottleneck of feature-representation-based methods. Although both DREAM and our work explore the gap between frontal-profile feature pairs, DREAM mostly follows empirical observations and makes approximate estimation for the differences between the feature pairs, while our approach theoretically captures and embeds the rotation relationship between the feature pairs. To the best of our knowledge, our work is the first attempt to explore how face rotation in the 3D space impacts on the deep feature  generation  process  of CNNs, and it is  critical to mathematically reveal the relationship between the rotation and resulting features.  Benefiting from the theoretical finding, LARNet has achieved superior performance, demonstrated by extensive experiments.


\section{Methodology}\label{sec:tp}
  In this section, we assume that a frontal face and its profile face have a corresponding rotation relationship in the original 3D space. For ease of understanding, only the rotation with the orthogonal transformation relationship  is discussed here. The derivation of more complex Euclidean transformation relationships, including translation and zooming, is referred to our supplementary material.  
 
 \subsection{Problem Formulation}
 Our goal is to find a transformation between the features of an input profile face image and the expected frontal face image,  to realize  \emph{frontalizition} in the deep feature space  and to achieve a powerful feature representation robust to pose variations, as shown in Fig.~\ref{fig:intro}.

 We denote $\mathcal{F}(\mathbf{x})$ as a feature extraction function in CNNs for an input image $\mathbf{x}$. Here, for each pixel $(u,v)$ in the image $\mathbf{x}$, we adopt its homogeneous coordinate representation 
 $(u,v,1)^{\top}$, and for convenience still denote the collection of these {3D} homogeneous coordinates as $\mathbf{x}$. 
 
 Let $d$ be the dimension of layers to be  considered. Then the extracted feature $\mathcal{F}(\mathbf{x}) \in  \mathbb{R}^{d}$. We shall prove that there exists a map $\mathcal{R}_{map}(\cdot) : \mathbb{R}^d \rightarrow \mathbb{R}^{d}$ that acts as a rotation in the deep feature space corresponding to the rotation  $\mathbf{R}\in SO(3)$ of the (homogenized) image $\mathbf{x}$ :
\begin{align}\label{eq:allrela}
  \mathcal{F} (\mathbf{R} \cdot \mathbf{x})= \mathcal{R}_{map}(  \mathcal{F}(\mathbf{x})).
 \end{align}
 
For the frontal face image $\mathbf{x}_f$ and its profile face image $\mathbf{x}_p$,  the homography transformation matrix of these two images degenerates into a rotation matrix:  $\mathbf{x}_f = \mathbf{R}\cdot \mathbf{x}_p $,  and we have:
\begin{align}\label{eq:fprela}
 \mathcal{F} (\mathbf{x}_f) = \mathcal{F}( \mathbf{R}\cdot \mathbf{x}_p) = \mathcal{R} _{map}(  \mathcal{F}(\mathbf{x}_p)).
\end{align}

Furthermore, we try to use  Lie group theory~\cite{Rossmann2002} and prove that the  mapping $\mathcal{R} _{map}(\cdot)$ can be decomposed into an additive residual component, which is solely determined by the rotation:
\begin{align}\label{eq:gcfrela}
 \mathcal{F} (\mathbf{x}_f) =  \mathcal{F}(\mathbf{x}_p) + \omega(\mathbf{R}) \cdot \mathbf{C}_{res} (\mathbf{R}, \mathbf{x}_p).
\end{align}
Thus, we only need a residual subnet $\mathbf{C}_{res}$ for decoding pose variant information from the input face image, and a gating subnet $\omega$ to learn rotation magnitude for controlling the strength of the residual component contributing to the feature learning process. 
Eq.~(\ref{eq:gcfrela}) is the core principle of LARNet we propose, and the detailed derivations and experimental design  will be presented in the following sections.

 \subsection{Rotation in Networks and Lie Algebra}
 To find what  $\mathcal{R} _{map}$ exactly is,  we have tried to directly explore and analyze the role of rotation $\mathbf{R}$ in networks from Eq.~(\ref{eq:fprela}).  From the original paper of ResNet~\cite{He2016}, it proposes a novel \emph{shortcut}, which not only retains the depth of deep networks, but also has the advantages of shallow networks in avoiding the overfitting issue. The feature learning from shallow layer $l$ to deep layer $L$ is described as:
  \begin{align}\label{eq:xl}
   \mathbf{x}_{L}= \mathbf{x}_{l}+\sum_{i=l}^{L-1} H\left(\mathbf{x}_{i}, w_{i}\right),
 \end{align}
 \begin{align}\label{eq:loss}
 \begin{split}
 \frac{\partial \mathbf{Loss}}{\partial \mathbf{x}_{l}} &= \frac{\partial \mathbf{Loss}} {\partial \mathbf{x}_{L}} \cdot \frac{\partial \mathbf{x}_{L}} {\partial \mathbf{x}_{l}}\\
 &=\frac{\partial \mathbf{Loss}}{\partial \mathbf{x}_{L}}\left(1+\frac{\partial}{\partial \mathbf{x}_{l}}\sum_{i=l}^{L-1} H\left(\mathbf{x}_{i}, w_{i}\right)\right),
 \end{split}
\end{align}
 where $\mathbf{x}_{l}$  represents the input of the $l$-th residual block, and $H(\cdot)$ is the residual function with weights $w$. Since the second term at big brackets of Eq.~(\ref{eq:loss}) will quickly drop to $1$, we focus on the first principal term $\partial \mathbf{Loss} / {\partial \mathbf{x}_{L}}$. 
 
Note that the rotation matrix $\mathbf{R}\in SO(3)$ is not closed under matrix additions. Hence in nonlinear optimization in CNNs, an update of $\mathbf{R}$ using derivations does not yield a new rotation matrix~\cite{Doug2015}. Therefore, a direct use of  $\mathbf{R}$ is not appropriate, while we need to explore a new approach of embedding $\mathbf{R}$ in the network. 

 
 Inspired by the prior work~\cite{Tuzel2008}, we adopt Lie algebra with its own addition, multiplication, and derivative to replace the rotation matrix $\mathbf{R}$ in CNNs. First, for each rotation matrix $\mathbf{R}\in \mathbb{R}^{3 \times 3}$, 
 it corresponds to a vector $\boldsymbol{\phi}$ through the exponential mapping~\cite{Carmo1992}:
 \begin{equation}
 \mathbf{R}=\exp(\boldsymbol{\phi}^{\wedge}),
 \label{eq:R}
 \end{equation}
 where $^{\wedge}$ is  the skew-symmetric operator. The detailed definition of  operator $^{\wedge}$ and a proof for Eq.~(\ref{eq:R})
 are placed into the supplementary material.
 
 On the other hand, the vector $\boldsymbol{\phi}$ can be obtained from $\mathbf{R}$ by the following Rodriguez' rotation formula \cite{Rodr1840} and Taylor expansion:
\begin{align}\label{eq:rform}
\begin{split}
\mathbf{R} &= \exp (\theta \psi^{\wedge})=\sum_{n=0}^{\infty} \frac{1}{n !}\left(\theta \psi^{\wedge}\right)^{n}\\
           &= \cos \theta \mathbf{I}+(1-\cos \theta) \psi\psi^{T}+\sin \theta \psi^{\wedge}. \\
\end{split}
\end{align}


Here $\boldsymbol{\phi}=\theta \psi$  is in the \emph{Axis-Angle representation} form,  with a unit vector $\psi \in \mathbb{R}^3$  being the direction of the rotation axis and $\theta$ being the rotation angle according to the right hand rule, respectively. Since $\mathbf{R}\psi=\psi$, $\psi$ is the eigenvector of the matrix $\mathbf{R}$ for eigenvalue $\lambda_{\mathbf{R}}=1$. Eq.~(\ref{eq:rform}) leads:
\begin{align}\label{eq:trance}
\begin{split}
tr(\mathbf{R})= 2 \cos \theta + 1.
\end{split}
\end{align}
Hence, we can solve $\boldsymbol{\phi}$ as:
\begin{align}\label{eq:solphi}
\begin{split}
\boldsymbol{\phi}= \theta \psi = \arccos( \frac{tr(\mathbf{R})-1}{2}) \psi.
\end{split}
\end{align}
Next, we show the addition and multiplication in Lie algebra by \emph{Baker-Campbell-Hausdorff} (BCH) formula~\cite{Wulf2002,Brian2015} and \emph{Friedrichs'} theorem~ \cite{Wilhelm1954,Nathan1966}:
\begin{align}\label{eq:addmul}
\begin{split}
&\exp \left(\Delta \boldsymbol{\phi}^{\wedge}\right) \exp \left(\boldsymbol{\phi}^{\wedge}\right)=\exp \left(\left(\boldsymbol{\phi}+\mathbf{J}_{l}(\boldsymbol{\phi})^{-1} \Delta \boldsymbol{\phi}\right)^{\wedge}\right),\\
&\exp \left((\boldsymbol{\phi}+\Delta \boldsymbol{\phi})^{\wedge}\right)=\exp \left(\left(\mathbf{J}_{l} \Delta \boldsymbol{\phi}\right)^{\wedge}\right) \exp \left(\boldsymbol{\phi}^{\wedge}\right).
\end{split}
\end{align}
 $\mathbf{J}_{\ell}$ is the $left$ $Jacobian$ of $SO(3)$. For a point $\mathbf{p} \in \mathbb{R}^3 $, the derivative of $\mathbf{R}\mathbf{p}$ with respect to a perturbed rotation is:
 \begin{align}\label{eq:der}
\begin{split}
  \frac{\partial(\mathbf{R}\mathbf{p}) }{\partial (\Delta\boldsymbol{\phi})}&=\lim _{\Delta\boldsymbol{\phi} \rightarrow 0} \frac{\exp \left(\Delta\boldsymbol{\phi}^{\wedge}\right) \exp \left(\boldsymbol{\phi}^{\wedge}\right)\mathbf{p} -\exp \left(\boldsymbol{\phi}^{\wedge}\right)\mathbf{p} }{\Delta\boldsymbol{\phi}}\\
    &=-(\mathbf{R} \mathbf{p})^{\wedge}.
\end{split}
\end{align}


For a current $\mathbf{R}_{i}$, we choose a perturbation $\Delta \phi^{\wedge}$, such that  $\mathbf{R}_{i+1}=\exp(\Delta\phi^{\wedge}) \mathbf{R}_{i}$.  Then for a point $\mathbf{p}$, Eq.~(\ref{eq:der}) leads:
\begin{align}
\label{eq:nonopt}
    \mathbf{R}_{i+1}\mathbf{p} = \exp (\Delta\boldsymbol{\phi}^{\wedge}) \mathbf{R}_{i} \mathbf{p}\approx \mathbf{R}_{i} \mathbf{p} - (\mathbf{R}_{i} \mathbf{p})^{\wedge} \Delta\boldsymbol{\phi}.
\end{align}
Then for the target function that is to be optimized, denoted by $u$, we use Taylor expansion to write:  
 \begin{align}\label{eq:uopt}
\begin{split}
 u(\mathbf{R}_{i+1}\mathbf{p})&=u\left(\exp \left(\Delta\boldsymbol{\phi}^{\wedge}\right) \mathbf{R}_{i} \mathbf{p}\right) \approx  u\left(\left(\mathbf{1}+\Delta\boldsymbol{\phi}^{\wedge}\right) \mathbf{R}_{i} \mathbf{p}\right) \\
              &\approx u\left(\mathbf{R}_{i} \mathbf{p}\right)-\underbrace{\left.\frac{\partial u}{\partial \mathbf{d}}\right|_{\mathbf{d} =\mathbf{R}_{i} \mathbf{p}}\left(\mathbf{R}_{i} \mathbf{p}\right)^{\wedge}}_{\boldsymbol{\delta}^{T}}\boldsymbol{\Delta\boldsymbol{\phi}} \\
             &=u\left(\mathbf{R}_{i} \mathbf{p}\right)+\boldsymbol{\delta}^{T} \boldsymbol{\Delta\boldsymbol{\phi}}.
\end{split}
\end{align}
We need to determine $\Delta\phi$ such that the value of $u$ decreases. A possible choice is to pick $\Delta\phi=-\alpha D \delta$, where $\alpha > 0$ is a small step size and $D$ is an arbitrary positive-definite matrix. Applying this perturbation within the scheme, we can update the rotation matrix by $\mathbf{R}_{i+1} \leftarrow \exp \left(-\alpha \mathrm{D} \delta^{\wedge}\right) \mathbf{R}_{i} $.

Back to the original problem, given Eqs.~(\ref{eq:addmul}-\ref{eq:uopt}), we can rewrite the first principal term of Eq.~(\ref{eq:loss}) as follows:
\begin{align}\label{eq:lossfp}
\begin{split}
 \frac{\partial \mathbf{Loss}}{\partial \mathbf{x}_L^f} &\approx \lim _{\Delta\boldsymbol{\phi}  \rightarrow 0} \frac{\partial\mathbf{Loss}} {\exp\left((\boldsymbol{\phi}+\Delta\boldsymbol{\phi})^{\wedge}\right) \cdot \mathbf{x}_L^p -\exp \left(\boldsymbol{\phi}^{\wedge}\right) \mathbf{x}_L^p} \\
                                        &=  \frac{\partial \mathbf{Loss}} {-(\mathbf{R}\cdot\mathbf{x}_L^p)^{\wedge} \cdot \partial \Delta\boldsymbol{\phi}}   \\
                                        &=  \frac{\partial \mathbf{Loss}} { \partial (\mathbf{R}\cdot\mathbf{x}_L^p)}.  
\end{split}                                        
\end{align}
Note that in Eq.~(\ref{eq:fprela}), we mentioned that the homography relationship between the two original images $\mathbf{x}_p$ and $\mathbf{x}_f$ is connected by a rotation, but this relationship generally cannot be guaranteed in the CNNs. However, Eq.~(\ref{eq:lossfp}) suggests that this relationship is inherited in another way during gradient decent at each layer. In fact, since $\mathbf{R} \in SO(3)$, $\mathbf{R}\cdot\mathbf{x}_L^p$ and $\mathbf{x}^f_L$ are asymptotically stable according to Lyapunov's second method~\cite{Lyapunov1992,Bhatia2002}. With the gradual training progress of the ResNet, the feature vectors of  $\mathbf{R}\cdot\mathbf{x}_L^p$ and $\mathbf{x}^f_L$ have the same convergent representation: $\mathcal{F}(\mathbf{x}_f) = \mathcal{F}(\mathbf{R}\cdot\mathbf{x}_p)$. Furthermore, we decouple the rotation relation from face features into Eq.~(\ref{eq:solphi})  and Eq.~(\ref{eq:nonopt}). Let $V_{res}=\mathcal{F}(\mathbf{R}\cdot\mathbf{x}_p)-\mathcal{R} _{map} \mathcal{F}(\mathbf{x}_p) \in \mathbb{R}^{d}$ be the residual vector, and we have:
\begin{align}\label{eq:xfvres}
\begin{split}
&\mathcal{R}^{-1} _{map} \mathcal{F}(\mathbf{x}_f) = \mathcal{F}(\mathbf{x}_p)+ \mathcal{R}^{-1} _{map} \cdot V_{res},  \\
&\mathcal{F}(\mathbf{x}_f)=\mathcal{F}(\mathbf{x}_p) + \mathcal{R}^{-1} _{map} (V_{res}+\mathcal{R}_{map}\mathcal{F}(\mathbf{x}_f)-\mathcal{F}(\mathbf{x}_f )).
\end{split}
\end{align}
Since during our training stage, the feature $\mathcal{F}(\mathbf{x}_p)$ is approaching to $\mathcal{R}_{map}\mathcal{F}(\mathbf{x}_f)$ (we shall show a corresponding analysis in Eq.~(\ref{eq:traning}) in Sec.~\ref{sec:subnet}), Eq.~(\ref{eq:xfvres}) leads to Eq.~(\ref{eq:xfxp}):
\begin{align}\label{eq:xfxp}
       \mathcal{F}(\mathbf{x}_f) \approx\mathcal{F}(\mathbf{x}_p) + \mathcal{R}^{-1} _{map} (\mathcal{F}(\mathbf{x}_p) - \mathcal{R}_{map}\mathcal{F}(\mathbf{x}_p)).
\end{align}
This agrees exactly with Eq.~(\ref{eq:gcfrela}). Hence, we can design the gating control function $\omega(\mathbf{R})$ as $\mathcal{R}^{-1} _{map}$ to filter the feature flow,  and  the component $\mathbf{C}_{res}(\mathbf{R}, \mathbf{x}_p) =\mathcal{F}(\mathbf{x}_p) - \mathcal{R}_{map}\mathcal{F}(\mathbf{x}_p)$ is obtained through residual network training. 

\subsection{The Architecture of Subnet}\label{sec:subnet}
As previously stated,  we expect to design a residual subnet $\mathbf{C}_{res}$ for decoding pose variant information from input face images.  The residual formulation in Eq.~(\ref{eq:gcfrela}) allows us to use a succinct enough network structure for learning the residual compensation from the clean deep features, which is a relatively easy task. The most convenient way  is to add the gating control residuals directly to the existing backbone (Arcface~\cite{Deng2019} with ResNet-50 in our paper). Residual learning can be arranged before the final fully-connected~(FC) layer of the ResNet-50 without revising any learned parameters of the original backbone model. Our residual learning has two fully-connected layers with Parametric Rectified Linear Unit (PReLU) ~\cite{He2015} as the activation function. We train it by minimizing $\ell_2$ norm of the difference between the profile features $\mathcal{F}(\mathbf{x}_p)$ and frontal features under the rotation $\mathcal{R}_{map}\mathcal{F}(\mathbf{x}_f)$ using stochastic gradient descent.
\begin{align}\label{eq:traning}
    \min\limits_{\Omega_{p}} \Sigma ||\mathcal{F}(\mathbf{x}_p) -   \mathcal{R}_{map}(\Omega_{p})\mathcal{F}(\mathbf{x}_f) ||_2^2,
\end{align}
where $\Omega_{p}$ denotes the learnable parameters.  We train this subnet on frontal-profile pairs sampled from the MS-Celeb-1M  dataset (mentioned in Sec.\ref{sec:dataset}), and fix these parameters for the testing. Applying a subnet with complicated structure may increase the risk of overfitting, and the design with two FC layers is on the consideration of both the task difficulty and the risk of model robustness. 

Furthermore,  we design a gating control function $\omega$ to analyze the rotation magnitude for controlling the strength of the residual component contributing to the deep feature learning process. In our problem context, $\omega$ needs to satisfy the following properties:

\begin{figure}[t]
 \centering
  \includegraphics[width=\linewidth]{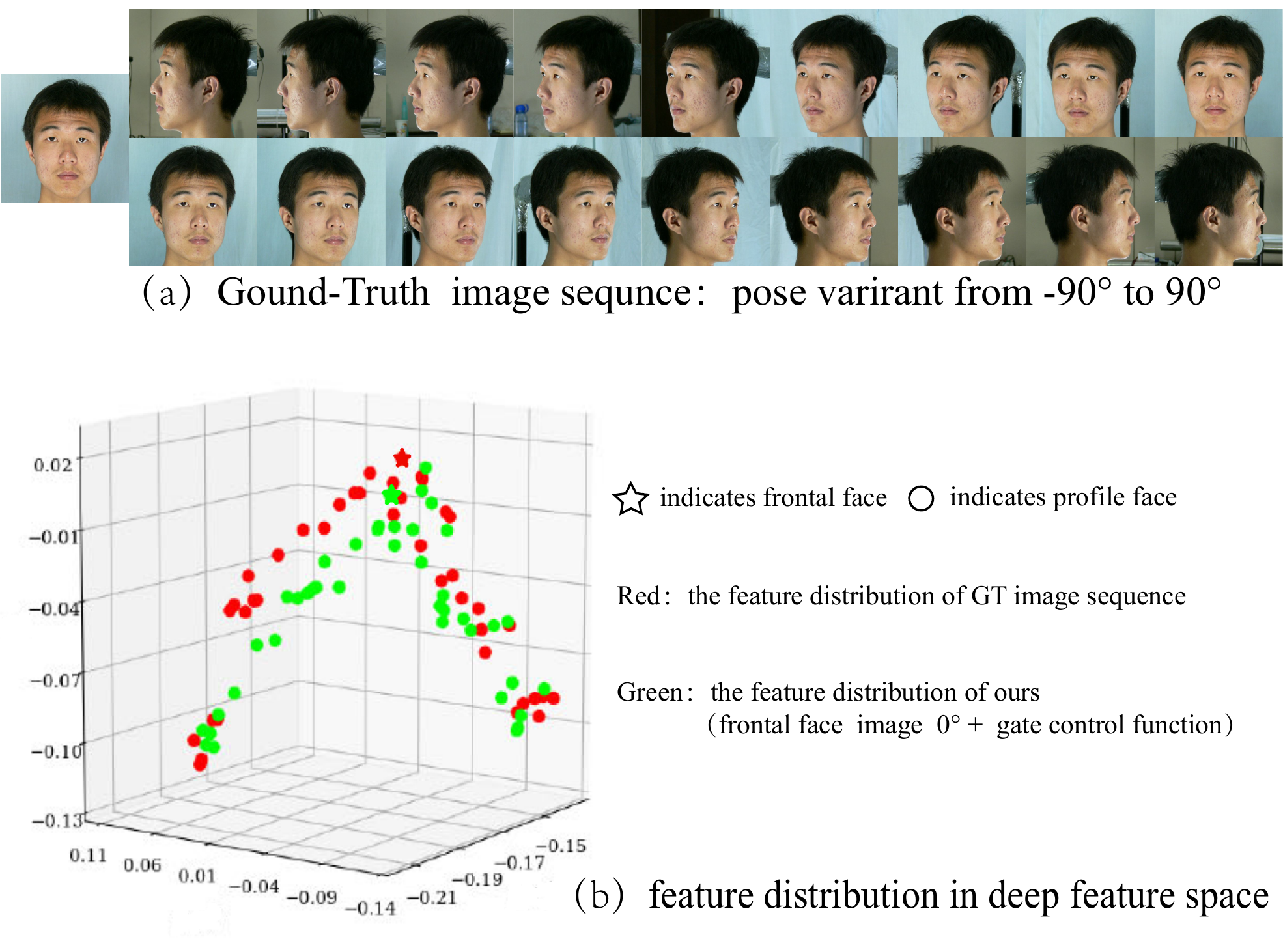}
  \vskip -3mm
  \caption{The effect of our gating control function for the same identity. (a) The top  is a sequence of images taken in real life, with pose variant being from $-90^\circ$ to $90^\circ$ for the same individual. (b) The bottom is the feature distribution in the deep feature space. The dots represent  profile faces while stars denote frontal faces. The red dots are the feature vectors generated by the image sequence, and the green dots are the feature vectors of the frontal face image ($0^\circ$) with different yaw angle variants simulated by our gating control function. Their similar distributions indicate that our gating control function maps the features of the frontal and  profile faces closer, thereby enhancing the feature representation ability to accommodate pose variations.}
  \label{fig:gcfsi}
  \vskip -3mm
\end{figure}

$\bullet$  $ \omega \in [0,1]$. Intuitively, when the input is frontal face $\mathbf{x}_0$, there is almost no difference in the feature representation in the same network, and  $\mathbf{C}_{res}$ of residual learning will bring errors and weaken the classification ability. Therefore, it is expected that the gating control function is 0 in this case; ideally, the magnitude of the residual is thus the largest at the complete profile pose: $ \mathcal{F}(\mathbf{x}_{0})- \mathcal{F}(\mathbf{x}_{\pi /2})$, so the maximum value of the gating control function is  1:
\begin{align*}
\mathcal{F}(\mathbf{x}_{0}) &= \mathcal{F}(\mathbf{x}_{ \pi /2}) + 1* (\mathcal{F}(\mathbf{x}_{0})- \mathcal{F}(\mathbf{x}_{\pi /2}))= \mathcal{F}(\mathbf{x}_{0}).
\end{align*}
$\bullet$  $ \omega $ has symmetric weights.  A gating control function learns rotation magnitude for controlling the strength of the residual component contributing to the feature learning process, and the same deflection angles should bring the same influence on frontal faces (for example, yaw angle with turning left or right). We will also use  data flipping augmentation to strengthen the symmetry of the model during training.

Besides, it is worth noting that roll, yaw, and pitch angles have different contributions to the final face recognition performance.  The effect of roll will be eliminated by face alignment (mentioned in Sec.~\ref{sec:dataset}), while face images  with large pitch angles are relatively rare. Combining all of the above constraints and solving Eq.~(\ref{eq:solphi}) by Chebyshev polynomial approximation, we get $\omega=|\sin{\theta}|$ with $\sin{\theta}=||(\sin_{pitch},\sin_{yaw},\sin_{roll})||_\infty$ for all angles $\in [-\pi/2,\pi/2]$, which ensures that there is a one-to-one correspondence between the elements in Lie algebra $\phi$ and the rotation $\mathbf{R}$ and  also guarantees the completeness of the proposed theory. 

\begin{figure}[t]
  \centering
  \includegraphics[width=\linewidth]{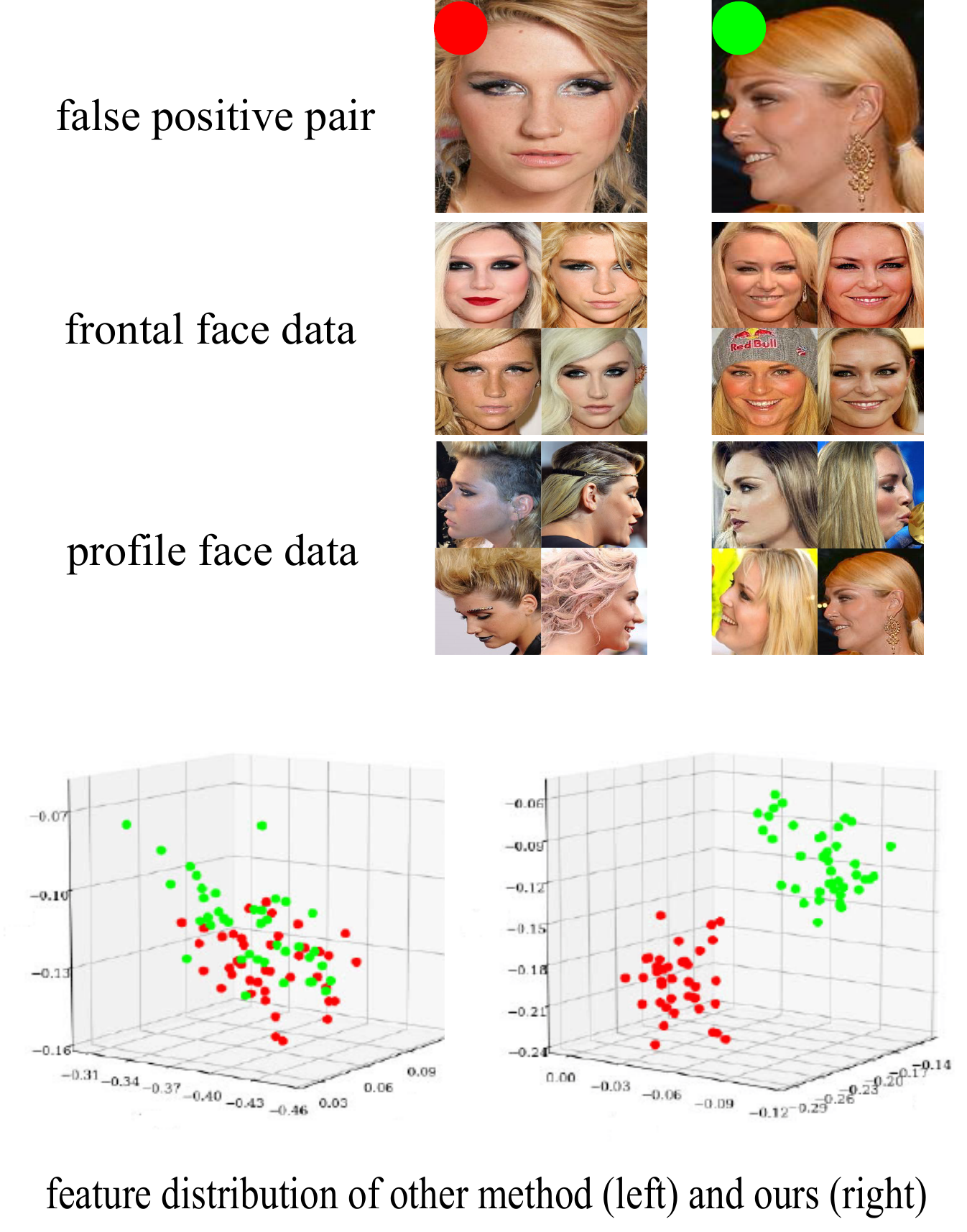}
  \vskip -3mm
  \caption{The effect of our gating control function for different identities. This is a challenging false positive example for a general face recognition model~\cite{Wang2018}. We collect more frontal and profile face data of those two individuals from the Celebrities Frontal-Profile dataset~\cite{Sengupta2016},  and visualize feature distributions of all the images. It is obvious that our model with the gating control function has a better classification and clustering ability.}
  \label{fig:gcfdi}
  \vskip -5mm
\end{figure}

As a way to demonstrate the effectiveness of our proposed subnet, Fig.~\ref{fig:gcfsi} illustrates for the same identity.  When inputs are image sequences of the same individual with different yaw angles, ResNet can extract features and display the distribution of those vectors as red dots. Meanwhile,  we use  our gating control function  with the only frontal face image to simulate pose variations. Our results' distribution is marked by green dots. Through the visualization, it can be clearly seen that our model can accurately simulate the feature vector distribution of different faces varying from yaw angles, which proves that our gating control function improves the feature representation capability, especially amenable to pose variations. 

In addition, Fig.~\ref{fig:gcfdi} illustrates the effect of our subnet for different identities.  This is a challenging example even for the almost blameless face recognition model~\cite{Wu2016}. We collect more frontal and profile face data of those two individuals, and visualize feature vectors of all the images  corresponding to this sample. It is obvious that our subnet with the gating control function is instrumental in obtaining better classification and clustering performance.


\section{Experimental Results}

In this section, we first provide a description of implementation details (Sec.~\ref{sec:ArchLAR}). Besides, we  list all the datasets used in the experiments and briefly explain their own characteristics (Sec.~\ref{sec:dataset}). Furthermore, we present two ablation studies on the architecture and gating control function, respectively, which explain the effectiveness of our experimental design on recognition performance (Sec.~\ref{sec:abs}). We also compare with existing methods and some findings about profile face representation,  and conduct extensive experiments on frontal-profile face verification-identification and general face recognition tasks (Sec.~\ref{sec:comex}).


\subsection{Implementation Details}\label{sec:ArchLAR}
\textbf{Network Architecture}. (1) Backbone: our backbone network is ResNet with a combined margin loss: CM(1, 0.3, 0.2) proposed by ArcFace. According to recent researches, ~\emph{i.e.}, Saxe~\emph{et al.}~\yrcite{Saxe2014}, Highway Networks~\cite{Srivastava2015}, and Balduzzi~\emph{et al.}~\yrcite{Balduzzi2017},  ResNet-50 has the best layers balancing efficiency and accuracy. Therefore, we choose the ResNet with 50 layers as it is very popular in a large amount of existing work and also convenient to compare. (2) LARNet: after the backbone, the clean deep features are mapped to \emph{rotation} through the subset and gating control function. (3) LARNet+: similar to DREAM~\cite{Cao2018}, we also use the end-to-end mechanism to further improve the performance of our results. Based on LARNet,  we make residual learning together with the backbone network in an end-to-end manner. We train the ResNet and residual learning together and then train the residual learning separately with pose-variant frontal-profile face pairs. 

\begin{figure}[t]
  \centering
  \includegraphics[width=\linewidth]{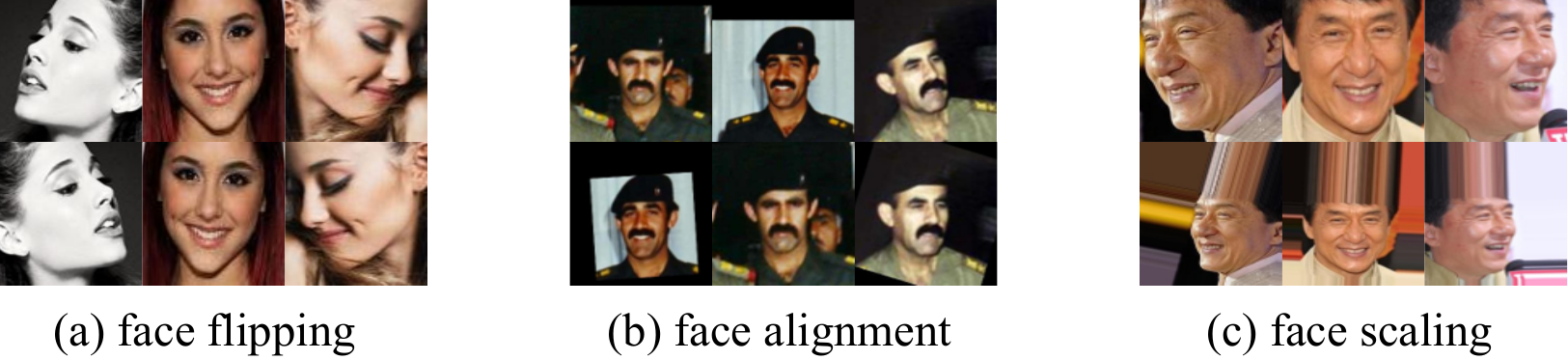}
  \vskip -3mm
  \caption{Data prepossessing on frontal and profile faces.  (a) Face flipping: data enhancement for strengthening the model's ability to learn symmetry; (b) face alignment: reduce the impact caused by translation and rotation in a plane such that the eyes lie along a horizontal line;  (c) face scaling: reduce the impact caused by zooming because focal length differs in every image  such that faces are approximately identical in size.    }
  \label{fig:fdpro}
  \vskip -5mm
\end{figure}

\textbf{Data Preprocessing}. As shown in Fig.~\ref{fig:fdpro},  we use MTCNN ~\cite{Zhang2016} to detect face areas and facial landmarks on both training and testing sets. We use a flipping strategy to achieve data augmentation and enhance our model's ability to learn symmetry. In addition,  face alignment and scaling ($224\times224$) are taken into account to reduce the impact caused by translation and zooming, when we only consider $SO(3)$ instead of $SE(3)$. 

\textbf{Training Details}. The model is trained  with 180K iterations.  The initial learning rate is set as 0.1, and  is divided by 10 after 100K, 160K iterations. The SGD optimizer has momentum $0.9$, and weight decay $5e{-4}$. For rotation angles, we take rotation estimation via state-of-the-art work~\cite{Yang2019, Yang2020} and obtain pose labels. The protocols of training and testing for different datasets will be explained in the next section.

\subsection{Datasets Exhibition}\label{sec:dataset}

\noindent{\textbf{Training Data}.} We separately employ the two most widely used face datasets as training data in order to conduct fair comparison with the other methods, \emph{i.e.}, cleaned MS-Celeb-1M database ~(\textbf{MS1MV2})~\cite{Guo2016}  and \textbf{CASIA-WebFace}~\cite{Yi2014}.  MS1MV2 is a clean version of the original MS-Celeb-1M face dataset that has too many mislabeled images, containing 5.8M images of 85,742 celebrities.   CASIA-WebFace uses tag-similarity clustering to remove noise of the data source,  containing 500K images of 100K celebrities from IMDb.

\noindent{\textbf{Testing Data}.} We explore several efficient face verification datasets for testing. Celebrities in Frontal-Profile ~(\textbf{CFP})~\cite{Sengupta2016} is a challenging frontal to profile face verification dataset, containing 500 celebrities, each of which has 10 frontal and 4 profile face images. We further test another challenging dataset IARPA Janus Benchmark A (\textbf{IJB-A})~\cite{Klare2015} that covers extreme poses and illuminations, containing 500 identities with  5,712 images and 20,414 frames extracted from videos. Besides focusing on frontal-profile face verification, we also conduct experiments on general face recognition datasets to verify that our method can reach the state of the art on general face recognition tasks.  Including the most widely used \textbf{LFW}~\cite{Huang2008} dataset (13,233 face images from 5,749 identities) and \textbf{YTF}~\cite{Wolf2011} dataset (3,425 videos of 1,595 different people), we also report the performance of Cross-Pose LFW (\textbf{CPLFW})~\cite{Zheng2018}, which deliberately searches and selects 3,000 positive face pairs with pose differences to add pose variations to intra-class variance,  so the effectiveness of several face verification methods can be fully justified. Furthermore, we also extensively conduct a more in-depth ablation experiment on the Large-scale CelebFaces Attributes (\textbf{CelebA}) dataset~\cite{Liu2015}, containing 10,177 celebrities and 202,599 face images, which covers large pose variations.

\begin{table}[t]
  \centering
  \vskip -3mm
  \caption{Ablation study on architecture. Evaluation is conducted on the CFP-FP dataset. }
 
    \resizebox{1.0\linewidth}{!}
  {
  \setlength{\tabcolsep}{0.1\linewidth}
  \begin{tabular}{|c|c|}
    \hline
    Architecture & Verification~(\%) \\
   \hline\hline
 Backbone  & 92.96  \\
 LARNet & 98.84  \\
 LARNet+  & \textbf{99.21} \\
\hline
  \end{tabular}
  }
  \label{table:abss}
  \vskip -3mm
  \end{table}

 \begin{table}[t]
  \centering
  \vskip -3mm
  \caption{Ablation study on gating control function. Evaluation is conducted on  the CelebA dataset with metric Equal Error Rate.}

  \resizebox{1.0\linewidth}{!}
  {
  \begin{tabular}{|l|c|}
   \hline
    Gating Control Function  &EER~(\%)  \\
   \hline\hline
Identity mapping:~$\omega\equiv 1$  & 15.35 \\
Linear mapping:~$\omega=2\theta/\pi$   & 9.68 \\
Nolinear mapping:~$\omega=sigmoid(4\theta / \pi-1)$  & 8.45 \\
PReLU & 9.72 \\
cReLU with OW & 7.92 \\
LARNet:~$\omega=|\sin{\theta}|$ & \textbf{6.26}\\
 \hline
  \end{tabular}
  }
  \label{table:gcf}
  \vskip -3mm
  \end{table}

\subsection{Ablation Studies}\label{sec:abs}
To justify that our LARNet does improve the performance of profile face recognition, we conduct two ablation experiments: 1) ~the architectures with/without  the gating control function, and 2) ~the forms of the gating control function.

 \begin{table*}[!htbp]
  \centering
  \vskip -3mm
  \setlength\tabcolsep{10pt}
  \caption{Quantitative evaluation on the IJB-A dataset, where o.s. denotes the optimal setting,  and f. is fine-tuning/refinement. Symbol ‘-’ indicates that the metric is not available for that protocol.}
 
     \resizebox{0.98\linewidth}{!}
  {
  \begin{tabular}{|l|cc|cc|}
    \hline
    Method	&TAR@FAR=0.01	&TAR@FAR=0.001	&Rank-1 Acc.	 &Rank-5 Acc.\\
    \hline\hline
 Wang~\emph{et al.} ~\yrcite{Wang2016}	&0.729	&0.510	&0.822	&0.931\\
 Pooling Faces~\cite{Hassner2016}	&0.819	&0.631	&0.846	&0.933\\
 Multi Pose-Aware~\cite{AbdA2016}	&0.787	&—	&0.846	&0.927\\
 DCNN Fusion (f.)~\cite{Chen2016}	&0.838	&—	&0.903	&0.965\\
 PAMs~\cite{Masi2016}	&0.826	&0.652	&0.840	&0.925\\
 Augmentation+Rendered~\cite{Masi2016b}	&0.886	&0.725	&0.906	&0.962\\
\hline
 Multi-task learning~\cite{Yin2017}	&0.787	&—	&0.858	&0.938\\
 TPE(f.)~\cite{Sanka2017}	&0.900	&0.813	&0.932 &—\\
 DR-GAN~\cite{Tran2017}	&0.831	&0.699	&0.901	&0.953\\
 FF-GAN~\cite{Yin2017b}	&0.852	&0.663	&0.902	&0.954\\
 NAN~\cite{Yang2017}	&0.921	&0.861	&0.938	&0.960\\
\hline
 Multicolumn~\cite{Xie2018}	&0.920	&—	&—	&—\\
 VGGFace2~\cite{Cao2018b} &0.904	&—	&—	&— \\
 Template Adaptation(f.) ~\cite{Cross2018}	&0.939	&—	&0.928	&—\\
 DREAM~\cite{Cao2018}	&0.872	&0.712	&0.915	&0.962\\
 DREAM(E2E+retrain,f.)~\cite{Cao2018}	&0.934	&0.836	&0.939	&0.960\\
\hline

 FTL with 60K parameters (o.s.)~\cite{Yin2019}	&0.864	&0.744	&0.893	&0.947\\
 PFEs~\cite{Shi2019}	&0.944	&—	&—	&—\\
\hline
 DebFace~\cite{Gong2020}	&0.902	&—	&—	&—\\
 Rotate-and-Render~\cite{Zhou2020}	&0.920 	&0.825	&—	&—\\
 HPDA~\cite{Wang2020}	&0.876 	&0.803	&0.84	&0.88\\
 CDA~\cite{Wang2020b}	&0.911 	&0.823	&0.936	&0.957\\
\hline
LARNet	&0.941	&0.842	&0.936	&0.968\\
LARNet+ 	&\textbf{0.951}	&\textbf{0.874}	&\textbf{0.949}	&\textbf{0.971}\\

\hline
  \end{tabular}
  }
 
  \label{table:IJBA}
  \vskip -3mm
 \end{table*} 
\subsubsection{The architecture of LARNet}
We study the effectiveness of architectures with and without the gating control function as well as an end-to-end optimization manner. All the results are using the same backbone ResNet-50, combined margin loss in Arcface~\cite{Deng2019}, and MS1MV2 training dataset,  and evaluation is conducted on the CFP-FP dataset.  The baseline method is Arcface without refinement. As shown in Table~\ref{table:abss}, we observe that compared with the strong baseline without the gating control function, our LARNet brings a significant advancement by $5.88\%$ in verification accuracy. LARNet+ with the end-to-end optimization manner also plays an important role in further performance improvement, and achieves a more superior result $99.21\%$ than LARNet $98.84\%$.

\subsubsection{The gating control function}
Next, we further analyze that which type of gating control function has a greater contribution to the performance. For facilitating a fair comparison, all methods take the same  CASIA-WebFace dataset and ResNet-50 backbone for training, and evaluation is conducted on CeleA with metric Equal Error Rate (EER). 
In Table~\ref{table:gcf},  identity mapping means $\omega\equiv 1 $, which represents some GAN-based work, ignoring the internal connection of frontal-profile face pairs, and only relying on the generator and the discriminator to produce results.  Linear mapping:~$\omega=2\theta/\pi$ is a natural attempt and meets the design constraints. Besides, our gating control function essentially acts as a filtering activation function, and we compare it with two widely used activation functions PReLU~\cite{He2015} and cReLU with OW~\cite{Balduzzi2017}. Nolinear mapping:~$\omega=sigmoid(4\theta / \pi-1)$ is reported by DREAM~\cite{Cao2018}. Our method~$\omega=|\sin{\theta}|$ still achieves the best result: EER $6.26\%$. This observation ascertains our design of exerting a higher degree of correction to a profile face, and has a better understanding about the effectiveness of our proposed LARNet for profile face recognition.

\subsection{Quantitative Evaluation Results}\label{sec:comex}
We compare our method with more than 30 competitors published in the recent five years with respect to different metrics, which include template based, GAN, residual learning, and three-dimensional reconstruction, aiming at handling various tasks such as face search, face recognition, face verification, large pose recognition, etc. All numerical statistics are the best results obtained from original quotation, cross-reference, and experimental reproduction.

 \subsubsection{ IJBA Dataset: verification and identification  with state of the arts}\label{sec:qeijba}
 
 In this experiment, we evaluate our method on the challenging benchmark IJBA that covers full pose variations and complies with the original standard protocol. The evaluation metrics include popular True Acceptance Rate (TAR) at False Acceptance Rate (FAR) of 0.01 and 0.001 on the verification task, and  Rank-1/Rank-5 recognition accuracy on the identification task. All methods employ the same  MS1MV2 dataset and ResNet-50 backbone for training.

 In Table~\ref{table:IJBA}, we compare with various state-of-the-art techniques. Our LARNet reaches 0.941~(TAR@FAR=0.01), and after refinement with end-to-end retraining,  LARNet+ achieves a  better performance with 0.951. They have surpassed the other existing methods by a large margin. Our method also brings significant improvement in a more challenging metric TAR@FAR=0.001, with the results of 0.842 and 0.874, respectively.  Furthermore,  for face identification, LARNet has an advantage in both Rank-1 Acc.~(0.936) and Rank-5 Acc.~(0.968), and LARNet+ pushes the result to a higher level: Rank-1 Acc.~(0.949) and Rank-5 Acc.~(0.971). Our method has achieved superior performance on both   recognition and verification tasks.

 \subsubsection{CFP-FP Dataset: profile face verification challenge}\label{sec:qecfp}
 We employ CFP-FP as the frontal profile face verification dataset with the protocol that the whole dataset is divided into 10 folds each containing 350 same and 350 not-same pairs of 50 individuals. All methods employ the same  MS1MV2 dataset and ResNet-50 backbone for training. From Table~\ref{table:CFP-FP}, we observe that the  face verification results of state-of-the-art face recognition models are basically at $94\%+$. It is worth noting that in 2020, a latest work named universal representation learning face (URFace)~\cite{Shi2020} has reached an astonishing $98.64\%$ under the training of  the MS1MV2 dataset and the auxiliary learning of a large number of modules, such as variation augmentation, confidence-aware identification loss, and multiple embeddings. However, the result of our LARNet has reached $98.84\%$, which outperforms almost all competitors. Regarding further advanced LARNet+, to our best knowledge,  the $99.21\%$ performance is the first to surpass the reported human-level performance ($98.92\%$) on the CFP-FP dataset.

    \begin{table}[t]
  \centering
  \vskip -3mm
  \caption{Quantitative evaluation on the CFP-FP dataset, where o.s. denotes the optimal setting, and f. is fine-tuning/refinement. }
 
   \resizebox{1.0\linewidth}{!}
  {
  \begin{tabular}{|l|c|}
    \hline
    Method & Verification~(\%) \\
   \hline\hline 
 SphereFace (o.s.+f.) & 94.17 \\
 CosFace (o.s.) & 94.40 \\
 ArcFace (o.s.+f.) & 94.04 \\
 URFace (all modules, MS1MV2, o.s.) &98.64\\
 Human-level  &98.92\\
   \hline
 LARNet & 98.84  \\
 LARNet+  & \textbf{99.21} \\
 \hline
  \end{tabular}
  }
  \label{table:CFP-FP}
\end{table}

   \begin{table}[t]
  \centering
  \caption{Quantitative evaluation on the general face recognition datasets: LFW, YTF, and CPLFW, where o.s. denotes the optimal setting,  and f. is fine-tuning/refinement. Symbol ‘-’ indicates that the metric is not available for that protocol. For fairness, the training network and dataset: ResNet-50 + CASIA-WebFace.  Symbol ‘*’ indicates the methods under different training designs. The evaluation metric is verification rate~(\%).}
  
  \resizebox{1.0\linewidth}{!}
  {
  \begin{tabular}{|l|ccc|}
   \hline
   Method	&LFW	&YTF	&CPLFW\\
   \hline\hline 
HUMAN-Individual	&97.27	&-	&81.21\\
HUMAN-Fusion	&\textbf{99.85}	&-	&85.24\\
DeepID~\cite{Sun2014}	&99.47	&93.20	&-\\
Deep Face~\cite{Taigman2014}	&97.35	&91.4	&-\\
VGG Face~\cite{Parkhi2015}	&98.95	&97.30	&90.57\\
FaceNet~\cite{Schroff2015}	&99.63	&95.10	&-\\
Baidu~\cite{Liu2015b}	&99.13	&–	&-\\
Center Loss~\cite{Wen2016}	&99.28	&94.9	&85.48\\
Range Loss~\cite{Zhang2017}	&99.52	&93.70	&-\\
Marginal Loss~\cite{Deng2017}	&99.48	&95.98	&-\\
SphereFace(o.s.)~\cite{Liu2017}	&99.42	&95.0	&81.4\\
SphereFace+(o.s.)	&99.47	&–	&90.30\\
CosFace(o.s.)~\cite{Wang2018}	&99.51	&96.1	&-\\
CosFace*(MS1MV2,R64, o.s.)	&99.73	&97.6	&-\\
Arcface(o.s.)~\cite{Deng2019}	&99.53	&–	&92.08\\
ArcFace*(MS1MV2,R100,f.)	&99.83	&\textbf{98.02}	&95.45\\
\hline
 Ours: LARNet 	&99.36	&96.55	&95.51\\
 Ours: LARNet+ 	&99.71	&97.63	&\textbf{96.23}\\
 \hline
  \end{tabular}
  }
  \label{table:generfr}
  \vskip -3mm
 \end{table}

 \subsubsection{LFW, YTF, and CPLFW Datasets: general face recognition}\label{sec:qegfr}
 To get a better understanding about our LARNet, we conduct a more in-depth comparison on general face recognition.  LFW and YTF datasets are the most widely used benchmark for unconstrained face verification on images and videos. In this experiment, we follow the unrestricted with labelled outside data protocol to report the performance. CPLFW emphasizes pose difference to further enlarge intra-class variance. All methods employ the same  CASIA-WebFace dataset and ResNet-50 backbone, while the results of some different experimental designs (marked with ‘*’) are also reported.

From Table~\ref{table:generfr}, we find that because the LFW dataset is too small, almost all methods can achieve $99\%+$. Although the meaning is very weak, the result $99.74\%$ of our method is also at the forefront, only lower than human-level $99.85\%$ and Arcface* $99.83\%$. It is worth noting that Arcface* is trained with the MS1MV2 dataset (5.8M) and ResNet-100, while Arcface with CASIA-WebFace (100K) and shallower ResNet-50 (the same as ours) only reaches $99.53\%$ and is inferior to ours. We provide this more comparative result for an extensive study. For the video sampled dataset YTF, there exists the same situation, and our LARNet and LARNet+ are still superior to all ResNet-50 based face recognition methods and slightly inferior to deeper ResNet-100 based Arcface*.  We also introduce a more challenging CPLFW dataset with large poses, which has more realistic consideration on pose intra-class variations. Our method achieves the best results $95.51\%$ and $96.23\%$, respectively.

\section{Conclusions}
We proposed the Lie Algebra Residual Network (LARNet) to boost face recognition performance. First, we presented a novel method with Lie algebra theory to explore how face rotation in the 3D space affects the deep feature generation process, and proved that a face rotation in  the image space is equivalent to an additive residual component in the deep feature space. Furthermore, we designed a gating control function, which is derived on the foundation of Lie algebra to learn rotation magnitude and  control the impact of the residual component on the feature learning process. Moreover, we provided the results of ablation studies to validate the effectiveness of our Lie algebraic deep feature learning. The comprehensive experimental evaluations demonstrate the superior performance of the proposed LARNet over the state-of-the-art methods on frontal-profile face verification, face identification, and general face recognition tasks. In future work, we will continue to pay attention to the interpretability of Euclidean transformation in other CNNs, and intend to explore more mathematical tools to decouple potential geometric properties hidden in the image space.

\section*{Acknowledgements}
This work was partially supported by the National Natural Science Foundation of China (Nos. 12022117, 61872354 and 61772523), the Beijing Natural Science Foundation (No. Z190004), and the Tencent AI Lab Rhino-Bird Focused Research Program (No. JR202023). 



\bibliography{example_paper}
\bibliographystyle{icml2021}

\clearpage
\setcounter{section}{0}
{\center{\LARGE{\textbf{Supplementary Material}}}}

\section{The existence of Lie algebra}
\textbf{Claim:} ~Given any rotation matrix $\mathbf{R} \in \mathbb{R}^{3\times3}$,  $\exists ~\boldsymbol{\phi} \in \mathbb{R}^{3}$ and a skew-symmetric operator $^{\wedge}$, there exists an exponential mapping:
\begin{align}
\mathbf{R}=\exp(\boldsymbol{\phi}^{\wedge}).
\end{align}

\textbf{Proof:} ~Without loss of generality,  $\mathbf{R}$ is orthogonal and $\mathbf{R} \in SO(3)$, and we have $ \mathbf{R} \cdot \mathbf{R} ^{\top} = \mathbf{I} $.   We assume that there is a continuous transformation from the frontal face to the profile face, so the time parameter $t$ is introduced and we derive:
\begin{align}
\begin{split}
    &\dot{\mathbf{R}}(t) \mathbf{R}(t)^{\top}+\mathbf{R}(t) \dot{\mathbf{R}}(t)^{\top}=0,\\
   & \dot{\mathbf{R}}(t) \mathbf{R}(t)^{\top}=-\mathbf{R}(t) \dot{\mathbf{R}}(t)^{\top}=-(\dot{\mathbf{R}}(t) \mathbf{R}(t)^{\top})^{\top}.
\end{split}
\end{align}
The above equation has a skew-symmetric form, denoted as:
\begin{align}
\begin{split}\label{eq:ode}
    &\dot{\mathbf{R}}(t) \mathbf{R}(t)^{\top}= \boldsymbol{\phi}(t) ^{\wedge},       \\
    &\dot{\mathbf{R}}(t)=\boldsymbol{\phi}(t) ^{\wedge} \cdot \mathbf{R}(t) ,     
\end{split}    
\end{align}
where $\boldsymbol{\phi}=(\boldsymbol{\phi}_1,\boldsymbol{\phi}_2,\boldsymbol{\phi}_3)^{\top} \in \mathbb{R}^{3}$ and  its skew-symmetric matrix is:
\begin{align}\label{eq:asm}
 \boldsymbol{\phi}^{\wedge} = 
\left[ \begin{array}{ccc}
0 & -\boldsymbol{\phi}_3 & \boldsymbol{\phi}_2\\
\boldsymbol{\phi}_3 & 0 & -\boldsymbol{\phi}_1\\
-\boldsymbol{\phi}_2 & \boldsymbol{\phi}_1 & 0
\end{array} 
\right ].
\end{align}
Therefore, taking the derivative once is equivalent to multiplying $\boldsymbol{\phi} ^{\wedge}$ on the left side.  And Eq.~(\ref{eq:ode}) is an ordinary differential equation with parameter variables:
\begin{align*}
 \frac{d\mathbf{R}(t)} {dt} = \boldsymbol{\phi}(t)^{\wedge} \cdot \mathbf{R}(t),
\end{align*}
\begin{align}\label{eq:odeint}
\frac{d \mathbf{R}(t)} {\mathbf{R}(t)} = \boldsymbol{\phi}(t)^{\wedge} \cdot dt.
\end{align}
Integrating both sides of Eq.~(\ref{eq:odeint}) leads to:
\begin{align}
\begin{split}
\int d \ln{\mathbf{R}(t)} &= \int  \boldsymbol{\phi}(t)^{\wedge} dt + C,\\
\mathbf{R}(t)&= \exp^{\int  \boldsymbol{\phi}(t)^{\wedge} dt + C} ,\\
\end{split}
\end{align}
where $C$ is a constant, determined by the initial value. Based on Riemann integral, we have
\begin{align}
\begin{split}
\int  \boldsymbol{\phi}(t)^{\wedge} dt  \approx  \sum\limits_{i=0}^{n-1} \boldsymbol{\phi}(t')^{\wedge}\cdot(t_{i+1}-t_{i}), \; t'\in [t_{i},t_{i+1}],\\
(0 = t_0 < t_1<t_2<...<t_{n-1}<t_n=t)\\
\end{split}
\end{align}
when there is a sufficiently small partition $T= \{t_i\}_{i=0}^{n}$  such that for any $\delta> 0$ and all $i$, there exists $|t_{i+1}-t_{i}|< \delta$, and then approximation $\approx$ can be equivalent $=$. In order to facilitate the calculation, we assume that the initial value $\mathbf{R}(0)= I $, and we have:
\begin{align}
\begin{split}
\mathbf{R}(t)&= \exp^{\sum\limits_{i=0}^{n-1}  \boldsymbol{\phi}(t')^{\wedge} \cdot(t_{i+1}-t_{i})} \\
&=\lim_{\Delta t_i \to 0} \prod\limits_{i=0}^{n-1} \exp^{\boldsymbol{\phi}(t')^{\wedge} \cdot\Delta t_i}\\
&=\prod\limits_{t=0}^{n-1} \exp^{\boldsymbol{\phi}(t')^{\wedge} dt'}.\\
\end{split}
\end{align}
For any $t$ of rotation $\mathbf{R}(t)$, there exists another sufficiently small partition $S=\{s_i\}_{i=0}^{n}$ such that 
\begin{align}
\begin{split}
\mathbf{R}(t)=\prod\limits_{i=0}^{n-1} \mathbf{R}(s),  \; s\in [s_{i},s_{i+1}].\\ 
(0 = s_0 < s_1<s_2<...<s_{n-1}<s_n=t)
\end{split}
\end{align}
Thus,  we construct a new partition $M=T\bigcap S = \{m_i\}_{i=0}^{n}$, and for all $m\in[m_{i},m_{i+1}]$ , $\Delta m_i=|m_{i+1}-m_{i}|\to 0$, we have:
\begin{align}
\mathbf{R}(m)=\exp(\boldsymbol{\phi}(m)^{\wedge}).
\end{align}
Referring to product integral theory~\cite{Antonin2007}, including its Lebesgue type \uppercase\expandafter{\romannumeral2}-geometric integral~\cite{Michael1972}, it provides a guarantee from discrete back to continuous: when $\Delta m_i\to0$,  we construct a new function $f(m)$ and let $(\exp(\boldsymbol{\phi}(m)^{\wedge})^{dm}=1+f(m)dm$.  Based on the properties of equivalent infinitesimal: $x\simeq \ln(1+x)$,  the following equation holds:$$\ln(\exp\boldsymbol{\phi}(m)^{\wedge}) dm = \ln(1+f(m)dm) =f(m)dm. $$
Furthermore , we have the following equation holding true:
\begin{align}
\begin{split}
\prod\limits_{i=0}^{n-1} \exp^{\boldsymbol{\phi}(m)^{\wedge} dm} &=\prod\limits_{i=0}^{n-1} (1+f(m)dm) \\
             &=\exp(\int_{0}^{\top} f(t)dt)\\
             &=\exp(\int_{0}^{\top} \ln(\exp^{\boldsymbol{\phi}(t')^{\wedge}}) dt')\\
             &=\exp(\int_{0}^{\top} \boldsymbol{\phi}(t')^{\wedge} dt').
\end{split}
\end{align}
Hence, for a fixed observation moment $t$, this equation holds:
\begin{align} 
\mathbf{R}(t) = \exp { (\boldsymbol{\phi}(t)^{\wedge} ) }  = \exp { (\int_{0}^{\top} \boldsymbol{\phi}(m)^{\wedge} dm ) }.
\end{align}
 \rightline{\qedsymbol}

\section{The properties of Lie algebra}
Every matrix Lie group is  smooth manifolds and has a corresponding Lie algebra. Lie algebra consists of a vector space $\mathbb{G}$ expanded on the number field $\mathbb{F}$ and a binary operator, which is called as Lie bracket $[\cdot,\cdot]$ and defined by the cross product $[X,Y]=X \times Y$ when $\mathbb{G} = \mathbb{R}^3$  \cite{Doug2015}. Now we only need to check that $\boldsymbol{\phi}$ satisfies the four basic properties of Lie algebra:
\begin{enumerate}
    \item Closure  
    $$[\boldsymbol{\phi}_{1}^{\wedge},\boldsymbol{\phi}_{2}^{\wedge}]= \boldsymbol{\phi}_{1}^{\wedge} \boldsymbol{\phi}_{2}^{\wedge}-\boldsymbol{\phi}_{2}^{\wedge}\boldsymbol{\phi}_{1}^{\wedge}=(\underbrace{\boldsymbol{\phi}_{1}^{\wedge} \boldsymbol{\phi}_{2}}_{\in \mathbb{R}^{3}})^{\wedge} \in \mathfrak{so}(3).$$
    \item Alternativity  
       $$[\boldsymbol{\phi}^{\wedge},\boldsymbol{\phi}^{\wedge}]=\boldsymbol{\phi}^{\wedge}\cdot\boldsymbol{\phi}^{\wedge}-\boldsymbol{\phi}^{\wedge}\cdot\boldsymbol{\phi}^{\wedge}=0 \in \mathfrak{so}(3).$$
    \item Jacobi identity can be verified by substituting and applying the definition of Lie bracket.
    \item Bilinearity follows directly from the fact that $(\cdot)^{\wedge}$ is a linear operator.
\end{enumerate}
 Informally, we will refer to $\mathfrak{so}(3)$ as the Lie algebra, although technically this is only the associated vector space.
 
 Furthermore, the derivative of rotation space is $\boldsymbol{\phi}$. From the above \textbf{Claim} and Eq.~(\ref{eq:ode}), we have $\dot{\mathbf{R}}(t)=\boldsymbol{\phi} ^{\wedge} \cdot \mathbf{R}(t)$. And also setting $t_0 = 0$ and $\mathbf{R}(t_0)=I$, we perform the first-order Taylor expansion:
\begin{align} 
    \mathbf{R} \approx \mathbf{R}\left(t_{0}\right)+\dot{\mathbf{R}}\left(t_{0}\right)\left(t-t_{0}\right)=\mathbf{I}+\boldsymbol{\phi}\left(t_{0}\right)^{\wedge} \cdot(t).
\end{align}
$\boldsymbol{\phi}$ reflects properties of the  derivative of $\mathbf{R}$. Mathematically, we call it on the tangent space near the origin of $SO(3)$. \qedsymbol

\section{The exponential mapping of Lie algebra}
 From the above section, we can clarify the composition of Lie algebra:
\begin{align}
 \mathfrak{so}(3)=\left\{\boldsymbol{\phi} \in \mathbb{R}^{3} | \boldsymbol{\Gamma}=\boldsymbol{\phi}^{\wedge} \in \mathbb{R}^{3 \times 3}\right\} . 
\end{align}
Let perform Taylor expansion on it, but we note that the expansion can only be solved when it converges, and that the result is still a matrix:
\begin{align}
\exp \left(\boldsymbol{\phi}^{\wedge}\right)=\sum_{n=0}^{\infty} \frac{1}{n !}\left(\boldsymbol{\phi}^{\wedge}\right)^{n}.
\end{align}
As mentioned in our main paper,  vector $\boldsymbol{\phi}$ can be denoted as $\boldsymbol{\phi}=\theta \psi$. Using its properties: (odd power) $\psi^{\wedge}\psi^{\wedge}\psi^{\wedge}=-\psi^{\wedge}$ and (even power) $\psi^{\wedge}\psi^{\wedge}=\psi\psi^{\top}-I$, we have:
\begin{align}
\begin{split}
\exp (\boldsymbol{\phi}^{\wedge}) &=\exp (\theta \psi^{\wedge})=\sum_{n=0}^{\infty} \frac{1}{n !}\left(\theta \psi^{\wedge}\right)^{n}\\
&=\cos \theta \mathbf{I}+(1-\cos \theta) \psi\psi^{\top}+\sin \theta \psi^{\wedge}.
\end{split}
\end{align}
This formula is exactly the same as Rodriguez' rotation. Therefore, we can consider solving the rotation vector through the trace of the matrix:
\begin{align}
\begin{split}
tr(\mathbf{R}) &= tr(\cos \theta \mathbf{I}+(1-\cos \theta) \psi\psi^{\top}+\sin \theta \psi^{\wedge}) \\
        &=  \cos \theta  tr(\mathbf{I}) + (1-\cos \theta) tr(\psi\psi^{\top}) +\sin \theta tr(\psi^{\wedge})\\
        &= 2 \cos \theta + 1.
\end{split}
\end{align}
By solving the above equation, it can be found that the exponential map is surjective. But there exist multiple $\mathfrak{so}(3)$ elements $\theta + 2 k \pi , k\in \mathbb{Z} $ corresponding to the same SO(3). If the rotation angle $\theta$ is fixed at $[-\pi /2 ,+\pi/ 2]$, then the elements in the Lie group and Lie algebra have a one-to-one correspondence (bijection). This means that our Lie algebra can completely replace rotation and will not produce adversarial examples. Besides, for $\mathbf{R}\psi=\psi$, $\psi$ is the eigenvector of $\mathbf{R}$, the corresponding eigenvalue is $\lambda=1$, and it is very convenient to solve $\boldsymbol{\phi}=\theta \psi$. \hfill \qedsymbol


\begin{figure}[t]
  \centering
  \includegraphics[width=0.6\linewidth]{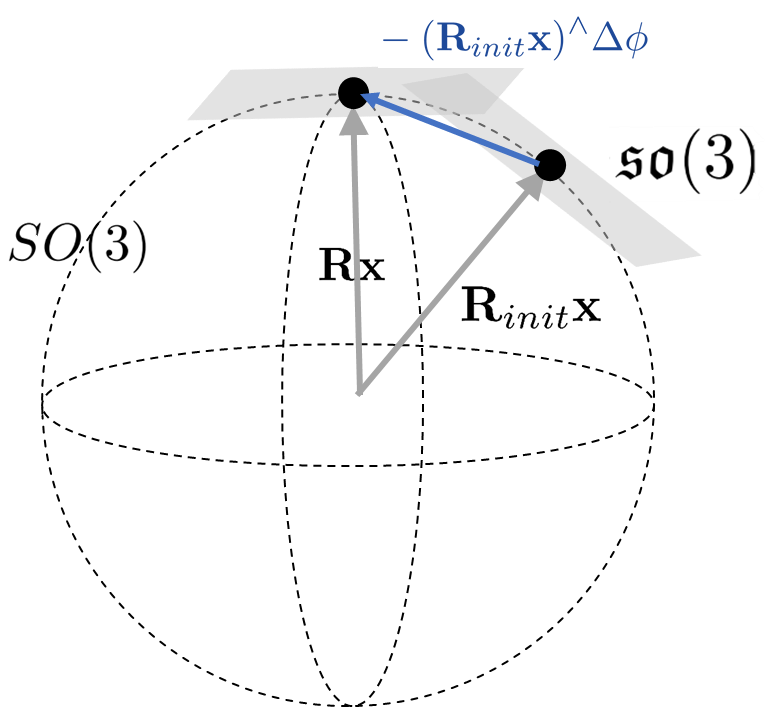}
  \caption{Nonlinear optimization of Lie algebra. During optimization, we keep our nominal rotation in the Lie group and consider a perturbation to take place in the Lie algebra, which is locally the tangent space of the group.}
  \label{fig:nonopt}
\end{figure}

\section{The optimization of Lie algebra}
 To compound two matrix exponentials, we use the \emph{Baker-Campbell-Hausdorff}~ (BCH) formula~\cite{Wulf2002,Brian2015} and \emph{Friedrichs'} theorem ~\cite{Nathan1966,Wilhelm1954} :
\begin{align}
\tiny
\ln (\exp (\mathbf{A}) \exp (\mathbf{B})) = \mathbf{A}+\sum_{n=0}^{\infty}(-1)^{n} \frac{B_{n}}{n !} \underbrace{[\mathbf{A},[\mathbf{A}, \ldots[\mathbf{A}, \mathbf{B}]}_{n} \ldots]],
\end{align}
where $B_n$ are $Bernoulli \ numbers$ and the Lie bracket is the usual $[\mathbf{A},\mathbf{B}]=\mathbf{A}\mathbf{B}-\mathbf{B}\mathbf{A}$ for $\mathbf{A},\mathbf{B} \in SO(3)$. 
Since the derivation can be seen as a change brought about by small increments, we have a more concise calculation model with vector representation using the approximate BCH :
\begin{align}
\ln \left(\exp \left(\boldsymbol{\phi}_1^{\wedge}\right) \exp \left(\boldsymbol{\phi}_2^{\wedge}\right)\right)^{\vee} \approx\left\{\begin{array}{ll}\mathbf{J}_{\ell}\left(\boldsymbol{\phi}_2\right)^{-1} \boldsymbol{\phi}_1+\boldsymbol{\phi}_2 \\ \quad\text { if } \boldsymbol{\phi}_1 \text { small } \\ \boldsymbol{\phi}_1+\mathbf{J}_{r}\left(\boldsymbol{\phi}_1\right)^{-1} \boldsymbol{\phi}_2 \\ \quad\text { if } \boldsymbol{\phi}_2 \text { small }\end{array} \right .    .
\end{align}
$\mathbf{J}_{\ell}$ and $\mathbf{J}_{r}$ are referred to as the $left$ and $right$ $Jacobians$ of $SO(3)$, respectively. Now for Lie algebra $\boldsymbol{\phi}$ and increment $\Delta\boldsymbol{\phi}$, we have:
\begin{align}
\begin{split}
&\exp \left(\Delta \boldsymbol{\phi}^{\wedge}\right) \exp \left(\boldsymbol{\phi}^{\wedge}\right)=\exp \left(\left(\boldsymbol{\phi}+\mathbf{J}_{l}(\boldsymbol{\phi})^{-1} \Delta \boldsymbol{\phi}\right)^{\wedge}\right),\\
&\exp \left((\boldsymbol{\phi}+\Delta \boldsymbol{\phi})^{\wedge}\right)=\exp \left(\left(\mathbf{J}_{l} \Delta \boldsymbol{\phi}\right)^{\wedge}\right) \exp \left(\boldsymbol{\phi}^{\wedge}\right)\\
&\qquad\qquad\qquad\quad\     =\exp \left(\boldsymbol{\phi}^{\wedge}\right) \exp \left(\left(\mathbf{J}_{r} \Delta \boldsymbol{\phi}\right)^{\wedge}\right)  .
\end{split}
\end{align}
Considering that the compound of rotation is left multiplication, we will work with the left increment and Jacobian. By comparing the derivative model on Lie algebra and the perturbation scheme on Lie group, we choose the latter for more conciseness, as follows:
\begin{align}
\begin{split}
    \frac{\partial\mathbf{R}\mathbf{x} }{\partial \Delta\boldsymbol{\phi}}&=\lim _{\Delta\boldsymbol{\phi} \rightarrow 0} \frac{\exp \left(\Delta\boldsymbol{\phi}^{\wedge}\right) \exp \left(\boldsymbol{\phi}^{\wedge}\right)\mathbf{x} -\exp \left(\boldsymbol{\phi}^{\wedge}\right)\mathbf{x} }{\Delta\boldsymbol{\phi}}\\
    &=-(\mathbf{R} \mathbf{x})^{\wedge},
\end{split}
\end{align}
where $\mathbf{x} \in \mathbb{R}^3 $ is an arbitrary three-dimensional point. When we take the product between rotation and a point with the perturbation scheme, we can get an approximation:
\begin{align}
    \mathbf{R}\mathbf{x} = \exp (\Delta\boldsymbol{\phi}^{\wedge}) \mathbf{R}_{init} \mathbf{x}\approx \mathbf{R}_{init} \mathbf{x} - (\mathbf{R}_{init} \mathbf{x})^{\wedge} \Delta\boldsymbol{\phi}.
\end{align}
This is depicted graphically in Fig.~\ref{fig:nonopt}. The above formula  has the form, which makes sense to nonlinear optimization like Gauss-Newton algorithm, and is adapted to work with the matrix Lie group by exploiting the surjective-only property of the exponential map. Furthermore, our scheme guarantees that it will iterate to convergence and $\mathbf{R}_{init} \in SO(3)$ at each iteration.  \hfill\qedsymbol

\section{Lie algebra of $SE(3)$}
We have proven that Euclidean transformation itself still has the same properties as $SO(3)$ with a more complex form, and we give its structure as:
\begin{align}
SE(3) = \left\{ \mathbf{T} = \left[ {\begin{array}{*{20}{c}} 	\mathbf{R} & \mathbf{t} \\ 	{{\mathbf{0}^{\top}}} & 1 	\end{array}} \right] \in \mathbb{R}^{4 \times 4} | \mathbf{R} \in SO(3), \mathbf{t} \in \mathbb{R}^{3} \right\}. 
\end{align}
Each transformation matrix has six degrees of freedom, so the corresponding Lie algebra is in $\mathbb{R}^6$, that is,
\begin{align}
\mathfrak{se}(3) = \left\{\boldsymbol{\xi}^\wedge \in \mathbb{R}^{4 \times 4} |~ \boldsymbol{\xi} \in \mathbb{R}^6 \right\},
\end{align}
where $^\wedge$ is not a skew-symmetric operator and has a new definition:
\begin{align}
\boldsymbol{\xi}^\wedge = {\left[ \begin{array}{l} 	\mathbf{\rho} \\ 	\mathbf{\phi}  	\end{array} \right]^ \wedge } = \left[ {\begin{array}{*{20}{c}} 	{{\mathbf{\phi} ^ \wedge }}&\mathbf{\rho} \\ 	{{\mathbf{0}^{\top}}}&0 	\end{array}} \right]  ,  ~	\mathbf{\phi}, \mathbf{\rho} \in \mathbb{R}^3.
\end{align}
Same as Eq.~(\ref{eq:ode}),   we also have an ordinary differential equation:
\begin{align}
    \mathbf{\dot{T}}(t) = \boldsymbol{\xi}^\wedge(t) \mathbf{T}(t).
\end{align}
Similarly, we can get its solution and the exponential mapping as:
\begin{align}
\begin{split}
    \mathbf{T}&=\exp \left(\boldsymbol{\xi}^{\wedge}\right)=\sum_{n=0}^{\infty} \frac{1}{n !}\left(\boldsymbol{\xi}^{\wedge}\right)^{n}\\
    &=\left[\begin{array}{cc}\sum_{n=0}^{\infty} \frac{1}{n !}\left(\phi^{\wedge}\right)^{n} & \left(\sum_{n=0}^{\infty} \frac{1}{(n+1) !}\left(\phi^{\wedge}\right)^{n}\right) \rho \\ 0^{\top} & 1\end{array}\right].
\end{split}
\end{align}
Similarly, we can define  addition operation, multiplication operation, and  the perturbation model. Because it has nothing to do with our work, we won't delve into details here.

Finally, we give  detailed calculation tables on the next page for Lie algebra, Lie group, and Jacobian. All our proofs and definitions are made up of the equations in the tables.
\begin{figure*}[t]
  \centering
  \includegraphics[width=0.98\linewidth]{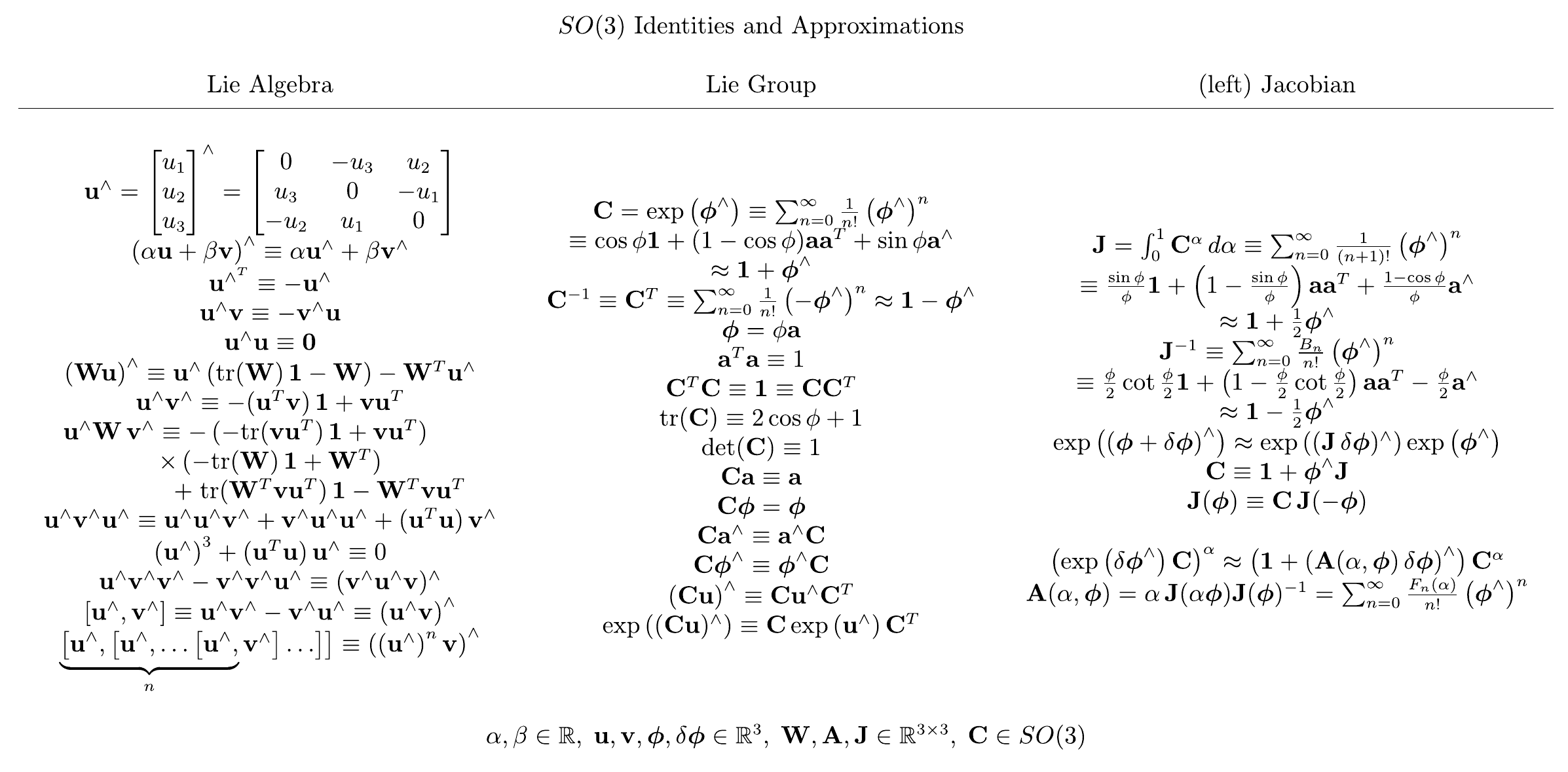}
  \vskip 10mm
\end{figure*}
\begin{figure*}[!h]
  \centering
  \includegraphics[width=0.98\linewidth]{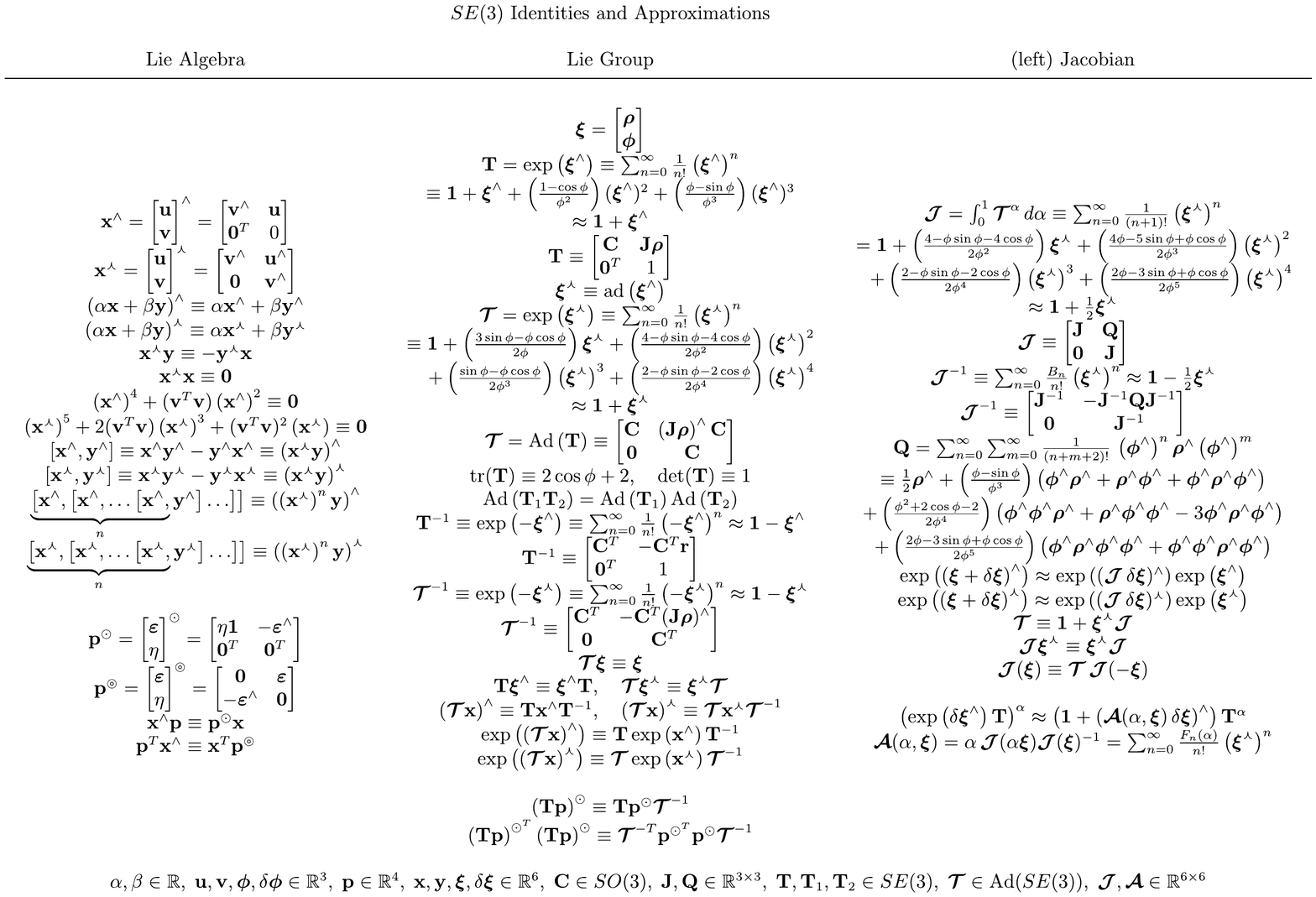}
   \vskip 10mm
\end{figure*}

\end{document}